\documentclass[conference]{IEEEtran} 
\pdfoutput=1
\usepackage{hyperref}
\usepackage{cite}
\usepackage{amsmath,amssymb,amsfonts}
\usepackage{algorithm,algorithmicx,algpseudocode}
\usepackage[inline]{enumitem}
\usepackage{booktabs}
\usepackage{graphicx}
\usepackage{textcomp}
\usepackage{xcolor}
\usepackage{subfig}
\def\BibTeX{{\rm B\kern-.05em{\sc i\kern-.025em b}\kern-.08em
    T\kern-.1667em\lower.7ex\hbox{E}\kern-.125emX}}

\DeclareMathOperator*{\argmin}{arg\,min}

\begin{document}
\graphicspath{{figures/}}

\title{A new heuristic algorithm for fast $k$-segmentation}

\author{\IEEEauthorblockN{1\textsuperscript{st} Sabarish Vadarevu}
\IEEEauthorblockA{\textit{Akridata India Pvt. Ltd.} \\
Bengaluru, India \\
sabarish.vadarevu@akridata.com}
\and
\IEEEauthorblockN{2\textsuperscript{nd} Vijay Karamcheti}
\IEEEauthorblockA{\textit{Akridata Inc.} \\
Los Altos, USA \\
vijay.karamcheti@akridata.com}
}

\maketitle

\begin{abstract}
    The $k$-segmentation of a video stream is used to partition it into $k$ piecewise-linear segments, so that each linear segment has a meaningful interpretation. Such segmentation may be used to summarize large videos using a small set of images, to identify anomalies within segments and change points between segments, and to select critical subsets for training machine learning models. Exact and approximate segmentation methods for $k$-segmentation exist in the literature. Each of these algorithms occupies a different spot in the trade-off between computational complexity and accuracy. A novel heuristic algorithm is proposed in this paper to improve upon existing methods. It is empirically found to provide accuracies competitive with exact methods at a fraction of the computational expense.
    
    The new algorithm is inspired by Lloyd's algorithm for K-Means and Lloyd-Max algorithm for scalar quantization, and is called the LM algorithm for convenience. It works by iteratively minimizing a cost function from any given initialisation; the commonly used $L_2$ cost is chosen in this paper. While the greedy minimization makes the algorithm sensitive to initialisation, the ability to converge from any initial guess to a local optimum allows the algorithm to be integrated into other existing algorithms. Three variants of the algorithm are tested over a large number of synthetic datasets, one being a standalone LM implementation, and two others that combine with existing algorithms. One of the latter two --- LM-enhanced-Bottom-Up segmentation --- is found to have the best accuracy and the lowest computational complexity among all algorithms. This variant of LM can provide $k$-segmentations over data sets with up to a million image frames within several seconds.
\end{abstract}


\section{Introduction}
\begin{figure*}
    \centering
   \includegraphics[width=0.7\linewidth]{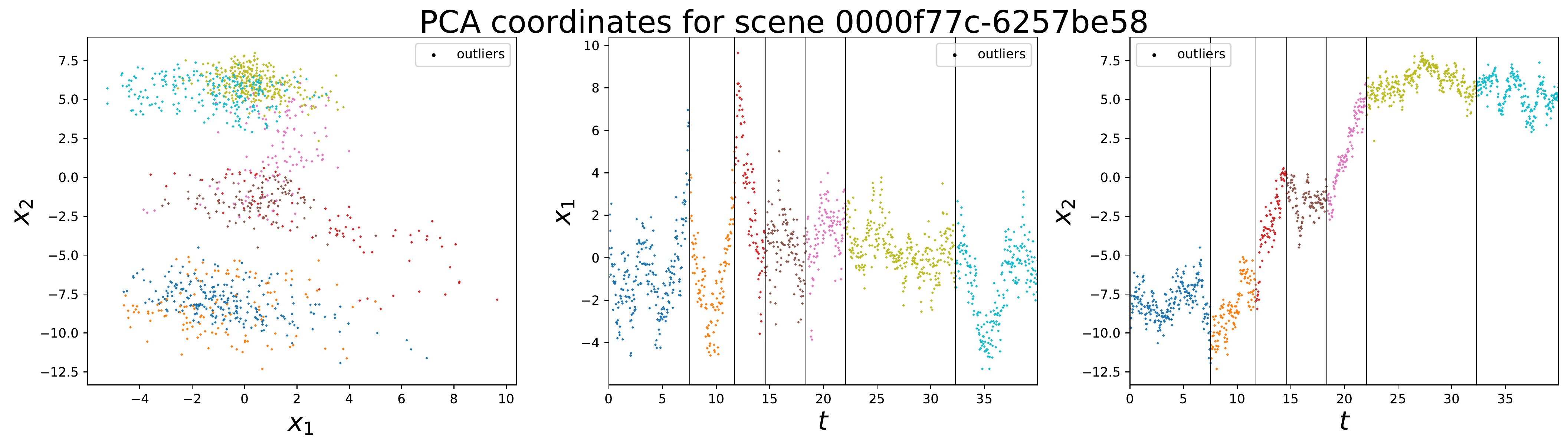}
    \caption{First two PCA coordinates (projections) for a 40s video from BDD100k; points in each segment are colored differently from their neighboring segments, and the ends are marked by solid vertical lines. Left shows first two PCA coordinates; center and right show first and second features against time.}
  \label{bdd-illustration-features} 
\end{figure*}

\begin{figure*} 
    \centering
    \includegraphics[width=0.85\linewidth]{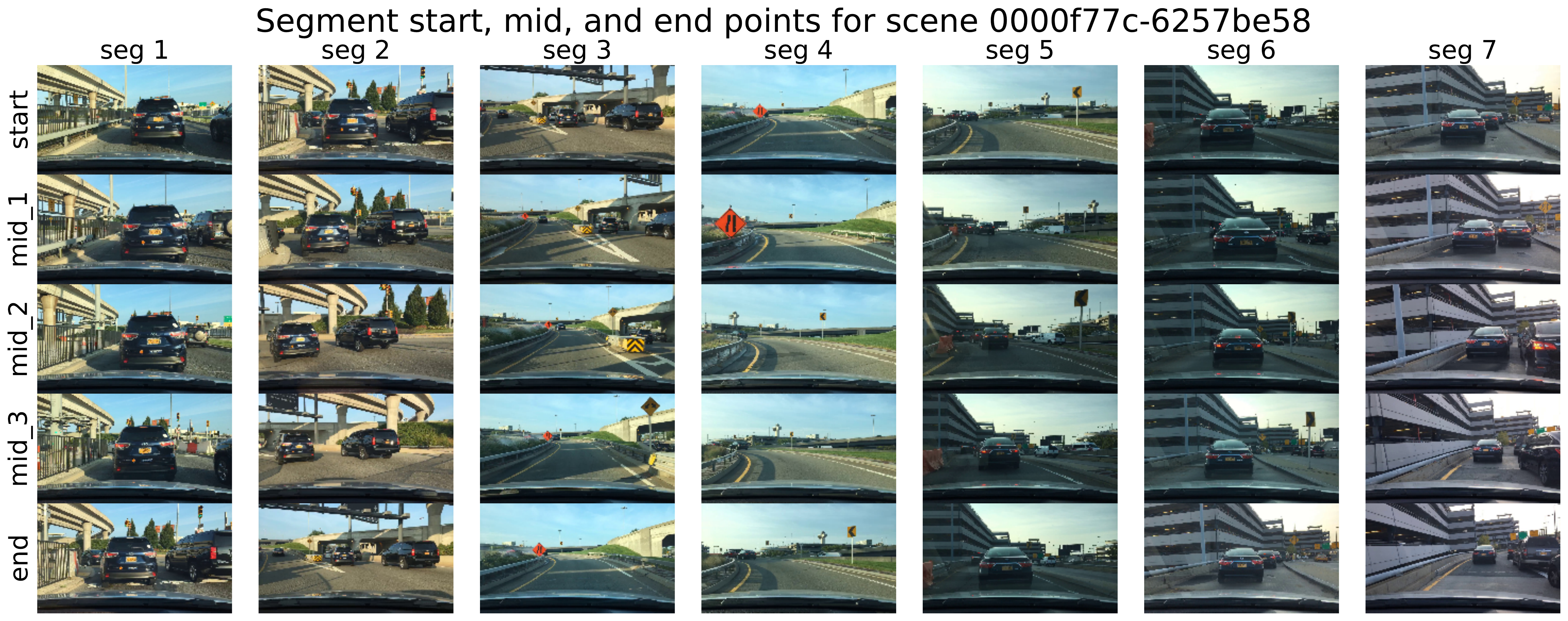}
    \caption{Optimal $7$-segmentation of a 40-second scene from BDD100k. Five images for are shown for each of the segments in Figure~\ref{bdd-illustration-features}, sampled uniformly in time within each segment.}
  \label{bdd-illustration-images} 
\end{figure*}

A simple example can motivate the $k$-segmentation problem. The Berkeley DeepDrive dataset (BDD100k) \cite{yu2020bdd100k} provides a large number of 40-second-long videos from diverse driving scenarios. One such video has been processed through a MobileNet featurizer \cite{sandler2018mobilenetv2}, and the first two PCA coordinates are plotted in Figure~\ref{bdd-illustration-features}. $K$-segmentation was performed ($k=7$) on these features, and the endpoints of the corresponding segments are marked. Five images corresponding to each segment are shown in Figure~\ref{bdd-illustration-images}. Each segment clearly corresponds to a particular driving regime. For instance, the seven segments can be described as follows.
\begin{enumerate}
    \item The recording car is slowly following behind another car.
    \item It slowly turns towards a different street as preceding cars speed up.
    \item It accelerates into the new street.
    \item It cruises along the new street.
    \item It slows down as it approaches a vehicle in front.
    \item It slowly follows the car in front.
    \item The car in front pulls away as the lane opens up.
\end{enumerate}
The PCA coordinates for the first, fourth, and sixth segments are nearly constant (with Gaussian-like noise), reflecting the uniform motion in the corresponding video segments. 

The meaningful partitioning provided by $k$-segmentation is not easily captured by other clustering algorithms. Such partitioning can be useful in several data science applications. For instance, a large video may be summarized using a small subset of images, either to reduce storage, reduce model training costs, or to simplify data exploration. Change points (transition points between segments) may be used to identify and correlate with non-trivial actions, such as lane-changes, turns, acceleration/deceleration, etc. To train a model targeted at identifying certain actions, say lane-changes, a few tagged images can be used to extract an entire segment of the video to provide the most relevant training data.

\subsection{Pre-processing}
For simplicity, a video is considered as just a time-ordered set of images, and all encoding/compression effects are ignored. A single video shall be considered a multi-dimensional signal (a signal is an indexed dataset), with each image in the video being considered a data point on the signal; the data points are ordered according to their time-stamp in the video.

\subsubsection{Image featurization}
Individual pixel values in images rarely contain information within themselves. In computer vision, the set of pixel values comprising an image are often channeled through a convolutional neural network (CNN) to produce a semantically rich set of features. Such features can then be used to perform classification, clustering, etc. In the context of  $k$-segmentation, each image in the data set is assumed to have been processed by some appropriate CNN, and the  $k$-segmentation itself is run on the resulting features. In the example shown above, the MobileNet model (checkpoint: mobilenet\_v2\_1.0\_224) \cite{sandler2018mobilenetv2} was used to extract 1280-dimensional features out of images. These features are taken as the activations at the final pooling layer in the CNN. MobileNet is used owing to its low computational expense, and the choice of checkpoint is arbitrary. Note that the time taken for $k$-segmentation itself, if the proper algorithm is chosen, can be much smaller than the time taken for image featurization.

While this paper motivates the $k$-segmentation problem as a means to semantically segment videos, the $k$-segmentation framework is also applicable to other domains where a signal is modelled to be piecewise-linear. The algorithm proposed in this paper is indeed relevant to other domains such as bioinformatics, economics, and epidemiology, among others.

\subsection{Overview}
This paper is organized as follows. In Section~\ref{sec:problem}, the $k$-segmentation problem is formally defined. Other variants of the problem are also introduced, along with a brief review of the literature on such problems. In Section~\ref{sec:algorithm}, a new heuristic algorithm is proposed to approximately solve the $k$-segmentation problem. Three variants of the new algorithm are introduced to demonstrate its adaptability. In Section~\ref{sec:results}, numerical experiments over a synthetic set of 500 multi-dimensional datasets are presented for the present algorithm along with several existing ones. The flexibility of the present algorithm becomes evident as it is used either as a standalone algorithm, or as an enhancement to existing methods. A real-world application of the algorithm is demonstrated in \cite{supplementary2020} for $7$-segmentations of 200 scenes from the Berkeley DeepDrive dataset \cite{yu2020bdd100k}; brief summary of these results is reported in Section~\ref{sec:bdd}.

\section{The $k$-segmentation problem}\label{sec:problem}
The $k$-segmentation problem defined below is specifically chosen to work with video data; other variants have been considered in the literature for change point detection. These differences are remarked upon in Section~\ref{sec:related-problems}.

\subsection{Notation}
\subsubsection{Time-indexed dataset}

Let $t_i$ be the time-index for the $i^{th}$ data point, with $t_i > t_j$ for $i>j$. Let $\mathbf{x}_i \in \mathbb{R}^d$ be the $d$-dimensional feature vector (produced by some CNN for an input image) at time $t_i$.
$X = \{ \mathbf{x}_1, \dots, \mathbf{x}_{N}\}$ is the dataset containing $N$ points indexed by 
times $T = \{t_1, \dots, t_{N}\}$, and
the set of tuples $P =\{(t_1, \mathbf{x}_1), \dots, (t_{N}, \mathbf{x}_{N})\}$ is the signal that we wish to compute $k$-segmentation over. The sub-signal over an interval $\lbrack t_{i+1},  t_j \rbrack$ is denoted $P_{i.j}$. The complete signal from $t_1$ to $t_N$ is $P_{0.N}=P$.

\subsubsection{$k$-segment}

For an integer $k \geq 1$, a $k$-segment is a $k$-piecewise-linear function $f_{k, i.j}: \mathbb{R} \to \mathbb{R}^d$ that maps each time $t_q$ in the interval $\lbrack t_{i+1}, t_j\rbrack$ to a $d$-dimensional vector $f_k(t_q) \in \mathbb{R}^d$. The $k$-segment $f_{k,i.j}$ is sometimes shortened to $f_k$ when the domain is clear from context. A $k$-segment is parametrized by (and equated to) a set of $k$ tuples,
$$f_{k, i.j} = \{(\mathbf{c}^1, \mathbf{m}^1, t_s^1=t_{i+1}, t_e^1), \dots,
(\mathbf{c}^{k}, \mathbf{m}^{k}, t_s^{k}, t_e^{k}=t_j)\},$$
where, $\mathbf{c}^j \in \mathbb{R}^d$ is the intercept for the $j^{th}$ segment, $\mathbf{m}^j \in \mathbb{R}^d$ is the slope of the $j^{th}$ segment, and $t_s^j$ and $t_e^j$ are the start and end points in time for the $j^{th}$ segment respectively. The end points $t_e^j$ are exclusive as a matter of convention, and obey
$ t_s^j < t_e^j = t_s^{j+1}.$

\subsubsection{Projection to $k$-segment}

The projection $f_k(t_i)$ of the indexed point $(t_i, \mathbf{x}_i)$ to the $k$-segment $f_k$ is calculated as follows:

\begin{itemize}
    \item Find the segment that contains $t_i$. That is, find $l: t_s^l \leq t_i < t_e^l$.
    \item The projection $f_k(t_i)$ is, using $l$ from above, $f_k(t_i) = \mathbf{c}^l + t_i ~ \mathbf{m}^l$.
\end{itemize}

\subsubsection{Fitting cost}

The fitting cost at time $t_i$ is the squared-distance for the $k$-segment-projection $f_k(t_i)$ from the associated point $\mathbf{x}_i$. The fitting cost for the sub-signal $P_{i.j}$ to the $k$-segment $f_{k, i.j}$ is
\begin{equation}\label{eq:cost}
    \text{cost}(P_{i.j}, f_{k, i.j}) = \sum\limits_{q=i+1}^{j} || \mathbf{x}_q - f_k(t_q) ||_2^{~2}.
\end{equation}
Euclidean distance (squared) is used above as the choice for fitting cost, but other metrics are possible.

\subsubsection{$k$-segment-mean}

A $k$-segment $f_{k,i.j}^*$ is the $k$-segment-mean of $P_{i.j}$ if $f_{k,i.j}^*$ minimizes the fitting cost, $\text{cost}(P_{i.j}, f_{k,i.j})$, over all possible $k$-segments on $\lbrack t_{i+1}, t_j\rbrack$.

\subsection{$k$-segmentation}
The $k$-segmentation problem is therefore that of finding the $k$-segment-mean for a given input signal, where the $k$-segment-mean is as defined above.

\subsection{1-segmentation}

For $k$=1, the $k$-segmentation algorithm degenerates to the case of linear regression. In this case, we seek a segment characterized by intercept $\mathbf{c}^*$ and slope $\mathbf{m}^*$ that minimizes the residual sum of squares (RSS) cost,
\begin{equation}
    (\mathbf{c}^*, \mathbf{m}^*) = \argmin\limits_{\mathbf{c}, \mathbf{m}} \sum\limits_{i=1}^{N} \big|\big|\mathbf{x}_i - (\mathbf{m} + t_i \mathbf{c}) \big|\big|_2^{~2}.
\end{equation}

The segment $(\mathbf{c}^*, \mathbf{m}^*, t_1, t_N)$ that minimizes the RSS cost is called the \textit{1-segment-mean} of the input signal $P_{0.N}$.

The fitting cost over the 1-segment mean of a sub-signal $P_{i.j}$ is simply denoted $\text{cost}(P_{i.j})$ without reference to the segment.

\subsection{Univariate regression}

Unlike usual regression problems in machine learning, the present $k$-segmentation problem has a single independent variable (time) and multiple dependent variables (image features). Because there is a single independent variable, the regression problem gets decoupled into a set of univariate regression problems. The intercept (bias) and slope (coefficients) are calculated for the sub-signal $P_{i.j}$ using covariances as
\begin{equation}
    \begin{aligned}
        \mathbf{m}^* &= \frac{\text{Cov}(T_{i.j}, X_{i.j})}{\text{Var}(T_{i.j})},\\
        \mathbf{c}^* &= \text{mean}(X_{i.j}) - \text{mean}(T_{i.j}) \cdot \mathbf{m}^* .
    \end{aligned}
\end{equation}

Here $\text{Cov}(T_{i.j}, X_{i.j})$ is the covariance of $T$ and each component of $\mathbf{x}_q \in X_{i.j}$; the covariances are not coupled to each other. $\text{Var}(T_{i.j})$ is the variance of time $T_{i.j}$.

\subsection{$k$-segmentation as a partitioning problem}

The $1$-segmentation problem is straight-forward as discussed above. The $k$-segmentation problem can be thought of being composed of two components:

\begin{enumerate}
    \item Decompose the dataset into $k$ optimal, contiguous (in time) partitions.
    \item Find optimal $1$-segmentation for each partition.
\end{enumerate}

Compared to the first sub-problem of partitioning, the second one is rather trivial. For a dataset with $N$ points, the number of possible combinations of $k$ partitions is 
${ N-2 \choose k-1} .$

In Section~\ref{sec:literature}, some of the popular algorithms for $k$-segmentation are presented, and a new heuristic algorithm is introduced in Section~\ref{sec:algorithm} to address some of the deficiencies.

\subsection{Related problems}\label{sec:related-problems}
The $k$-segmentation problem, as defined above, can be considered a special case of the more general problem of change point detection. In change point detection, each point $\mathbf{x}_i$ in a signal $\{(t_i, \mathbf{x}_i): i \in \{1, 2, \dots, N\}\}$ is considered to be drawn from some distribution $\mathfrak{f}_i(.)$, which is possibly parametrized. The problem is to find a set of points $\mathcal{T} = \{\tau_1, \tau_2, \dots, \tau_k\}$ so that each partition of points $\{\mathbf{x}_i: \tau_j \leq t_i < \tau_{j+1} \}$ contains points all drawn from the same distribution $\mathfrak{f}_j(\cdot)$ \cite{chen2000parametric}.

Parametric methods \cite{chen2000parametric} model the density function as a function of some parameter $\theta$, $\mathfrak{f}(\cdot|\theta)$, where $\theta$ can be multi-dimensional. A change point is said to occur at $\tau_j$ when the density distribution for the two sets $\{\mathbf{x}_i: \tau_{j-1} \leq t_i < \tau_{j}\}$ and $\{\mathbf{x}_i: \tau_{j} \leq t_i < \tau_{j+1}\}$ are parametrized by different $\theta_j$ and $\theta_{j+1}$ to some statistical significance. Non-parametric methods follow a similar approach, except that a statistically significant difference is required in cumulative density functions instead of the parameters of the probability density \cite{zou2014nonparametric}. Change point detection problems can be further divided into online and offline problems \cite{aminikhanghahi2017survey}. In this paper, we deal primarily with offline methods, with some brief remarks on online methods towards the end.

\subsubsection{The present $k$-segmentation problem}
The $k$-segmentation problem considered in this paper falls into the category of parametric offline change point detection methods. A recent review of such problems, along with others, was presented by \cite{truong2020selective}. Within this category of problems, two subcategories can be differentiated: problems with piecewise-constant signals, and problems with piecewise-linear signals. In either case, problems can be univariate or multivariate, and supervised or unsupervised. The problems can also have a known or an unknown number of change points.

In this paper, we are concerned with the unsupervised problem of multi-dimensional piecewise-linear signals with a known number of change points. In the typology of change point detection problems proposed by \cite{truong2020selective}, our $k$-segmentation problem has a fixed $L_2$ cost function and a known number of change points; the question is that of finding the most suitable search method. We further assume that the noise about the piecewise-linear model is Gaussian, and that the variance of the noise remains unchanged amongst segments. The same unsupervised problem can also be modeled using autoregression; we choose instead to use simple linear regression with time as the sole explanatory variable to simplify the 1-segmentations.

In the remainder of this paper, $k$-segmentation and change point detection are used interchangeably to refer to the current problem.

\subsection{Existing algorithms for $k$-segmentation}\label{sec:literature}
The $k$-segmentation problem has received a lot of interest in domains such as DNA sequencing and economics. Several algorithms have been proposed under different cost functions and constraints \cite{chen2000parametric,zou2014nonparametric,aminikhanghahi2017survey,truong2020selective,van2020evaluation}. In this paper, only six of the more popular algorithms are discussed. Some of these algorithms are exact but expensive, while the others are approximate and cheap. A detailed review of these algorithms, including pseudocode for implementation, is found in \cite{truong2020selective}.

\subsubsection{Note on complexity}
Computational complexity of change point detection algorithms is often reported in terms of the number of cost computations \cite{truong2020selective}; because these algorithms serve as search methods over a variety of cost functions. This paper is concerned with a single cost function, defined in (\ref{eq:cost}), whose time complexity is $\mathcal{O}(Nd)$. The time complexities reported in this paper include this complexity due to the cost function to provide a more accurate estimate. Space complexity is often ignored because, for the present offline $k$-segmentation problem, this is usually comparable or smaller to the space required to hold the signal to be segmented.

\subsection{Approximate $k$-segmentation}
A recent review \cite{truong2020selective} describes and provides pseudo-code for three approximate methods: Window Sliding (WS), Binary Segmentation (BS), and Bottom-Up segmentation (BotUp). For the sake of brevity, only BotUp is outlined here; because a modification to BotUp is used later in this paper. The time complexity of these algorithms is mentioned in Table~\ref{tab:algo-complexity}.

Beyond the above three methods, an enhanced version of BS called Wild Binary Segmentation (WBS) \cite{fryzlewicz2014wild} was also proposed. For each sub-signal $P_{i.j}$, WBS considers a specified number of sub-signals of $P_{i.j}$ that are randomly sampled. A change point is detected by maximizing the cost discrepancy (difference in fitting costs between optimal $2$-segmentation and $1$-segmentation) over all of the randomly sampled sub-signals. WBS is not included in the numerical experiments in this paper because a Python implementation is not readily available, unlike BS, WS, and BotUp, which are provided by \cite{truong2020selective}. 

\subsubsection{Bottom-Up Segmentation (BU)}
Bottom-up segmentation begins by slicing the signal $P_{0.N}$ into a large number of ($m \approx N/\delta \gg k$) of uniformly-sized sub-signals, with $\delta$ typically 2 or larger; larger values make the resulting segmentation less accurate. Two adjacent sub-signals $P_{i.j},~P_{j.l}$ are then merged into one sub-signal $P_{i.l}$ when, over all pairs of adjacent cells, their merge produces the smallest fitting cost discrepancy \cite{truong2020selective}. This is continued until the required number $k$ of segments is reached.

\begin{table}[!t]
\renewcommand{\arraystretch}{1.3}
\caption{Time complexity of $k$-segmentation algorithms,\\
    including the complexity of each of the 1-segmentations.}
\label{tab:algo-complexity}
\centering
\begin{tabular}{|c|c||c|c|}
\hline
    \bfseries Approx. method & \bfseries Complexity & \bfseries Exact method & \bfseries Complexity\\
\hline\hline
    WS$^a$ & $Nwd$ & SN & $N^3kd$\\
    BS & $N^2d\log(N)$ & SNBC\_sf50  & $Nd^2 + (30k)^3d$ \\
    LM\_(q)inits & $Nqd$ & SNBC\_sf10& $Nd^2 + (10k)^3d$ \\
    BotUp & $Nd$ & PELT  & $N^2d$ to $N^3d$ \\
    LM-BotUp & $Nd$ & SNBC\_sf10-LM & $Nd^2 + (10k)^3d$\\
\hline
\multicolumn{4}{l}{$^{\mathrm{a}}$For a sliding window of size $w$.}
\end{tabular}
\end{table}

\subsection{Exact $k$-segmentation}
Exact $k$-segmentation algorithms find globally optimal change points. The brute force approach would be to minimize cost over all possible ${N-2 \choose k-1}$ combinations. More efficient algorithms have been proposed that use dynamic programming to reduce the number of possible candidates for change points. A recent review \cite{truong2020selective} describes and provides pseudo-code for two exact methods --- Segment Neighborhood (SN) \cite{auger1989algorithms} and Pruned Exact Linear Time (PELT) \cite{killick2012optimal} --- along with a brief note on other extensions and enhancements. A third method that uses balanced coresets to reduce time complexity is outlined below. The time complexities for these algorithms is given in Table~\ref{tab:algo-complexity}.

Other exact algorithms have been proposed \cite{guedon2013exploring,rigaill2015pruned,haynes2017computationally}; these are variations or extensions of the above three algorithms, and will not be discussed in this paper.

\subsubsection{Segment Neighborhood over Balanced Coresets (SNBC) \cite{rosman2014coresets}}
SN limits the number of computations, but implicitly accounts for all ${N-2 \choose k-1}$ possible change point combinations. SNBC (name used by present authors for convenience) uses theoretical considerations to directly shrink the pool of candidates for change points \cite{rosman2014coresets}. The signal is sliced into a large number of disjoint and complete sub-signals under the constraint that considering only the start and end points of these sub-signals can provide solutions within a prescribed error bound. This algorithm is classified here as an exact method because the error bound can be made arbitrarily small.

A balanced subpartitioning step is used to come up with the set of disjoint and complete sub-signals. This is done by assigning a cost budget, and then accumulating sub-signals until the 1-segmentation cost for each sub-signal remains just below the cost budget. If the size of these sub-signals is large ($\gg d$), a $(k,\epsilon)$-coreset \cite{rosman2014coresets} is constructed so that repeated 1-segmentations in the SN algorithm become cheaper to compute.

The acceptable worst-case error is controlled by the cost budget used for balanced subpartitioning. Larger cost budgets reduce computational complexity, but may produce larger errors. The present authors use a hyperparameter `sf' (short for sigma-factor) to assign this cost budget $\sigma$ as
\begin{equation*}
    \sigma = \frac{\mbox{cost}_{\tiny\mbox{bicriteria-approximation}}}{k\cdot \mbox{sf}}
\end{equation*}
The original paper recommends sf~$\approx 100 \log_2(N)$. However, using such a large sf takes too long for practical implementations. Numerical experiments by the present authors (not shown here) on synthetic datasets with around 50,000 points showed that a cost budget with sf=50 that produces approximately $30k$ balanced coresets produces an acceptable trade-off between accuracy and time complexity. This version of SNBC is labelled SNBC\_sf50 in this paper; an even coarser implementation with sf=10 is also used here, and is labelled SNBC\_sf10.

\section{A new heuristic algorithm (LM)}\label{sec:algorithm}
The approximate methods reviewed previously look for one change point at a time. Furthermore, this is done by $\mathcal{O}(N)$ searches to account for every possible change point. A new heuristic algorithm is now described to address both of these issues. It is inspired by Lloyd's algorithm for $k$-means and Lloyd-Max algorithm for scalar quantization, and is therefore referred to as the LM algorithm in the remainder of this paper.

The LM algorithm is outlined in Algorithm~\ref{algo:LM}. It is modeled after the two-step refinement of cluster memberships and model parameters used in $k$-means and Gaussian mixture models. The analogue of cluster membership re-assignment is due to operations~\ref{algo:LM-repart}, \ref{algo:LM-repart1} and \ref{algo:LM-repart2}, where change points between segments are re-assigned after a local minimization. The analogue of model parameter modification is operation~\ref{algo:LM-reseg-1}. Both of these steps reduce the fitting cost; therefore, the fitting cost decreases monotonically as the iterations progress. The cost does not decrease anymore as the local infimum is approached. The stopping criterion in operation~\ref{algo:LM-stop} breaks the loop when the cost decrease becomes small enough.

LM retains some features of BS and WBS, such as finding a single, locally optimal change point within a sub-signal $P_{i.j}$, in operation~\ref{algo:LM-repart}. Like WBS, LM allows exploring different sub-signals via the initial $k$-segment $f^0_k$ that is provided as input. However, it differs from WBS in two key aspects. \begin{enumerate*}[label=(\roman*)]
    \item While each point in the signal is evaluated as a change point candidate in LM too, this is done by computing fitting cost over 1-segment-means from the previous iteration in operations~\ref{algo:LM-ptwise-1} and~\ref{algo:LM-ptwise-2} ; thus, the number of 1-segmentations is significantly reduced.
    \item The loop in operation~\ref{algo:LM-loop-segments} optimizes the change point within a single pair of segments. However, this optimization of change points is allowed to propagate to other pairs of segments in the same iteration.
\end{enumerate*}

\begin{algorithm}[ht!]
 \caption{Lloyd-Max-like heuristic algorithm for $k$-segmentation (LM)}
 \label{algo:LM}
 \begin{algorithmic}[1]
 \newcommand{\algorithmicbreak}{\textbf{break}}
 \newcommand{\BREAK}{\State \algorithmicbreak}
 \Require A signal $P_{0.N}$, number of segments $k$, max iterations $r_{\max}$, initial $k$-segment $f^0_k$, minimum segment size $\gamma$, and convergence tolerance $\epsilon$
 \Ensure Locally optimal $k$-segment $f^+_k$ 
  \State $\mathfrak{c}^0 \gets \mbox{cost}(P_{0.N}, f^0_k).$
     \Comment{Track fitting cost}
  \For {$r = 1$ to $r_{\max}$}\Comment{Improve $k$-segment iteratively}
    \State $Q \gets \{(g_{i,l.q}, g_{i+1,q.m}): g_i~\mbox{is the }i^{th} \mbox{ segment in}~f_{k}^{r-1}, i<k\}.$
    \For {$(g_i, g_{i+1}) \in \mbox{shuffled}(Q) $}\label{algo:LM-loop-segments}
  	  \State $P_{l.m}$ is the sub-signal covered by $g_i \cup g_{i+1}$
	  \For {$(t_q, \mathbf{x}_q) \in P_{l.m}$}
        \State $c^i_{q} \gets ||\mathbf{x}_q - g_i(t_q)||_2^{~2}.$
          \Comment{Pointwise cost} \label{algo:LM-ptwise-1}
        \State $c^{i+1}_{q} \gets ||\mathbf{x}_q - g_{i+1}(t_q)||_2^{~2}.$ \label{algo:LM-ptwise-2}
	    \EndFor
      \State $s^+ \gets \argmin\limits_{s = l+\gamma}^{m-\gamma} \bigg(
        \sum\limits_{q=l+1}^{s} c^i_q + \sum\limits_{q=s+1}^{m} c^{i+1}_q\bigg)$ \label{algo:LM-repart}
        \Statex \Comment{Re-compute change point} 
      \State Update end-point of $g_i$ to $s^+$ \label{algo:LM-repart1}
      \State Update start-point of $g_{i+1}$ to $s^+ + 1$ \label{algo:LM-repart2}
  \EndFor
  \State $f^{r-1/2}_k \gets \{g_i, ~i \in \lbrack 1, k\rbrack\}$ \Comment{Intermediate $k$-segment}
     \For {$g_{i,l.m} \in  f^{r-1/2}_k$}
      \State $g_{i,l.m} \gets \mbox{1-segment-mean}(P_{l.m})$ \label{algo:LM-reseg-1}
        \Statex \Comment{Re-compute and update 1-segment-mean}
  \EndFor
    \State $f^{r}_k \gets \{g_i, ~i \in \lbrack 1, k\rbrack\}$ \Comment{Updated $k$-segment}
    \State $\mathfrak{c}^r \gets \mbox{cost}(P_{0.N}, f^r_k)$
    \If {($\mathfrak{c}^r \geq (1-\epsilon)\cdot \mathfrak{c}^{r-1} $)}
      \BREAK \label{algo:LM-stop}
      \Comment{Stop if cost changes too slowly}
    \EndIf
  \EndFor
 \State $f^+_k \gets f^r_k$.
     \Comment{Locally optimal $k$-segment}
 \State \Return $f^+_k$
 \end{algorithmic}
 \end{algorithm}

The number of 1-segmentations required for LM is $\mathcal{O}(kr_{\max})$, where $r_{\max}$ is the maximum number of iterations. Numerical experiments over a large variety of multidimensional signals has shown that the iterations typically converge within a few iterations. Therefore, the number of 1-segmentations remains $\mathcal{O}(k)$. However, if the complexity of computing pointwise fitting costs in operations~\ref{algo:LM-ptwise-1} and~\ref{algo:LM-ptwise-2}, and of computing 1-segment-means in operation~\ref{algo:LM-reseg-1} are included, the time complexity of LM becomes $\mathcal{O}(Nd)$. By comparison, BS and WBS have complexity $\mathcal{O}(N^2d \log(N))$ when 1-segmentation costs are included.

\subsection{Variants of LM}
The LM algorithm is a flexible method that takes an initial set of candidate change points and produces a locally optimal set of change points. The simplest application of this algorithm involves starting with a uniformly spaced set of candidate change points and converging to the closest minimizing set. Because of the greedy nature of the algorithm, this often produces sub-optimal sets of change points. Three variants of LM are now proposed to make better use of the algorithm.

\subsubsection{LM\_(q)inits}
A set of $q ~(\sim 10)$ random initialisations are allowed to converge to the corresponding locally optimal $k$-segmentations. The converged solution with the smallest fitting cost is chosen as the final solution. In this paper, the performance of LM\_20inits (with 20 random initialisations) is reported.

\subsubsection{LM-BotUp}
LM is used to provide an accelerated initialisation for BotUp. BotUp starts with a very large number of cells ($\sim N/\delta$, for some small integer $\delta$), and successively merges these sub-signals. The LM-initialised-BotUp uses LM to produce a ``good'' initialisation with a significantly smaller number of cells, thereby reducing the number of merging operations required. In addition to reducing the time complexity, LM-BotUp also addresses the problem of premature discretization of BotUp, where initial assignment of cell boundaries and early merges can lead to the disappearance of the true change points from later consideration. In Section~\ref{sec:results}, this variant uses a $k_{\mathrm{init}} = \min(5k, N/20)$ number of segments for the LM component, with $k_{\mathrm{init}}$ uniformly sized segments for initialisation. The converged solution for this $k_{\mathrm{init}}$-segmentation is then used by BotUp to produce a $k$-segmentation.

\subsubsection{SNBC\_sf10-LM}
This variant incorporates a coarse implementation of SNBC: SNBC\_sf10. In SNBC\_sf10, the cost budget is significantly increased beyond the recommended values to produce fewer balanced coresets; this reduces time complexity at the expense of accuracy. This coarse solution is expected to identify change points that are sufficiently close to the globally optimal set, which can in turn be refined by LM to produce the true global optimum.

The time complexities of all the algorithms mentioned thus far are tabulated in Table~\ref{tab:algo-complexity}. All of the approximate methods except BS have a time complexity linear in the signal size $N$. While the scaling is linear for several of these, the proportionality constant describing the upper bound can be significantly different. For SN and PELT, the scaling is either quadratic or cubic in signal size. For variants of SNBC, the scaling is linear in signal size, but there is an additional $k$-dependent term with different weights for different sf values.

In the next section, comparative results for the above variants of LM against the existing algorithms are reported for a large number of numerical experiments.

\section{Evaluation of $k$-segmentation algorithms}\label{sec:results}
The algorithms used for comparison are Window Sliding (WS), Binary Segmentation (BS), Bottom-Up segmentation (BotUp), Segment Neighborhood (SN), Segment Neighborhood over Balanced Coresets (SNBC\_sf50, SNBC\_sf10), and Pruned Exact Linear Time (PELT). A Python package called \textit{Ruptures} is readily available \cite{truong2020selective} to run $k$-segmentation over the algorithms WS, BS, BotUp, SN, and PELT. For SNBC\_sf50 and SNBC\_sf10, a custom Python implementation is used. For WS, a window size of 50 is used, and the algorithm is labelled WS\_w50.

PELT controls the number of change points through a penalty parameter $\beta$. For consistency, different values of $\beta$ were tried, and the one that produces the required number of change points is used to produce the final solution. The run-times reported below are for the final run using the most appropriate penalty value.

\subsection{Evaluation metrics}\label{sec:metrics}
Several evaluation metrics have been used for change point detection \cite{truong2020selective,van2020evaluation}. We use covering metric and Rand index to evaluate the accuracy of the above-mentioned algorithms for the $k$-segmentation problem with known number of change points.

\subsubsection{Covering metric}
The set of true change points is denoted $\mathcal{T}^*$, and the set of predicted change points is denoted $\mathcal{T}^+$. The segmentation covering metric is a weighted average of the Jaccard indices for segment partitions $\mathcal{G}^*$ and $\mathcal{G}^+$ defined respectively over $\mathcal{T}^*$ and $\mathcal{T}^+$ as $\mathcal{G} = \{ \{t_q : \tau_i \leq t_q < \tau_{i+1} \}: \tau_i, \tau_{i+1} \in \mathcal{T} \}$; these partitions are the times between successive change points. The Jaccard index is the intersection over union measure for each partition,
\begin{equation}
    \mathcal{J}(\mathcal{A}^*, \mathcal{A}^+) = \frac{|\mathcal{A}^* \cap \mathcal{A}^+|}{|\mathcal{A}^* \cup \mathcal{A}^+|}, ~~~ \mathcal{A}^* \in \mathcal{G}^*,~ \mathcal{A}^+ \in \mathcal{G}^+,
\end{equation}
which are weighted by partition size to define the covering metric,
\begin{equation}\label{eq:covering}
    \Delta_C(\mathcal{G}^*, \mathcal{G}^+) = \frac{1}{|T|} \sum\limits_{\mathcal{A} \in \mathcal{G}}
                |\mathcal{A}|\cdot \max\limits_{\mathcal{A}^+ \in \mathcal{G}^+}
                \mathcal{J}(\mathcal{A}^*, \mathcal{A}^+).
\end{equation}

The covering metric provides an accuracy in the clustering sense of $k$-segmentation by looking for how well similar points are grouped together. The F1-score is used by \cite{van2020evaluation} to evaluate change points in the classification sense; however, for the current problem of known number of change points, the F1-score degenerates to the classification accuracy.

\subsubsection{Rand index}
Rand index measures the extent to which points from the same true segment are grouped together in the predicted segmentation, \textbf{and} the extent to which points from different true segments are grouped separately in the predicted segmentation \cite{truong2020selective}. The covering metric, in contrast, only measures the former of the two. For each set of change points, $\mathcal{T}^*$ and $\mathcal{T}^+$, two sets of tuples for grouped indices gr$(\mathcal{T})$ and non-grouped indices ngr$(\mathcal{T})$ are defined as
\begin{equation}
    \begin{aligned}
        \mbox{gr}(\mathcal{T}) = \{& (s,t),~ s,t \in T :\\
                                   &\exists ~ \tau_i, \tau_{i+1} \in \mathcal{T}
                                    \mbox{ with }  \tau_{i} \leq s, t < \tau_{i+1} \} \\
        \mbox{ngr}(\mathcal{T}) = \{& (s,t),~ s,t \in T :\\
                                   &\nexists ~ \tau_i, \tau_{i+1} \in \mathcal{T}
                                    \mbox{ with }  \tau_{i} \leq s, t < \tau_{i+1} \} 
    \end{aligned}
\end{equation}

The Rand index is then defined as 
\begin{equation}\label{eq:rand-index}
    \Delta_R(\mathcal{T}^*, \mathcal{T}^+) = \frac{ |\mbox{gr}(\mathcal{T}^*) \cap \mbox{gr}(\mathcal{T}^+)| + 
                      |\mbox{ngr}(\mathcal{T}^*) \cap \mbox{ngr}(\mathcal{T}^+)|}{ |T|\cdot (|T|-1) }.
\end{equation}

\subsection{Synthetic data}
Synthetic signals with known change point locations are generated to imitate real-world signals as in Figure ~\ref{bdd-illustration-features} to benchmark the performances of different algorithms. These signals are piece-wise linear and have different signal sizes $N$, with dimensionality $d \in \lbrack 2, 16\rbrack$. A weak non-linearity is introduced to the piece-wise linear base segments, along with three different kinds of noise --- Gaussian, high frequency trigonometric, and impulsive --- to reflect real-world conditions.

\begin{figure} 
    \centering
    \includegraphics[width=0.95\linewidth]{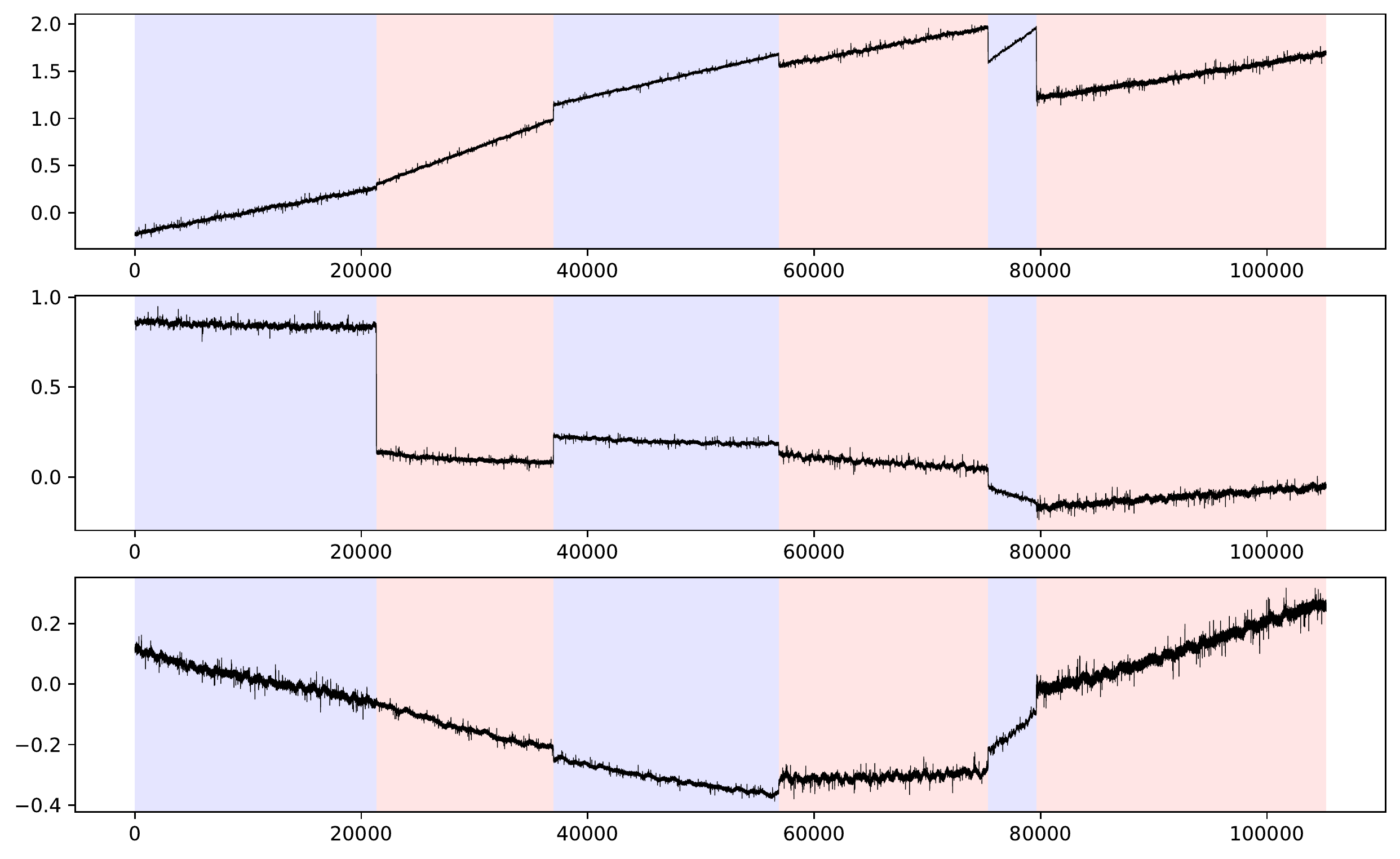}
    \caption{A synthetic signal, plotted against the location-index of each point in the signal. The first three PCA coordinates are shown from top to bottom. The alternating blue-red shading delineates neighboring segments.}
  \label{fig:example-data} 
\end{figure}

The non-linearities are introduced as second to fourth degree polynomials in time with their corresponding coefficients being significantly smaller than the coefficients for the constant and linear terms. These coefficients and noise levels are randomly chosen. An example signal is shown in Figure~\ref{fig:example-data}, with the segments shaded alternatively in blue and red.

\subsection{Detecting a single change point}
The problem of detecting a single change point serves as a starting point to evaluate $k$-segmentation algorithms. This problem is also critical in the context of online learning. For such problems, exact $k$-segmentation methods --- SN, PELT, and SNBC\_sf50 --- are not required, and BS can find the optimal change point through a single sweep of the entire signal. A set of 200 synthetic signals \textit{containing exactly two segments} are generated to evaluate the approximate methods: BS, BotUp, WS\_w50, along with two variants of the newly proposed LM algorithm, LM\_20inits and LM-BotUp. These signals have size $N \in \lbrack 400, 15000\rbrack$ and dimensionality $d \in \lbrack 2, 16\rbrack$.

\begin{figure} 
    \centering
    \includegraphics[width=0.95\linewidth]{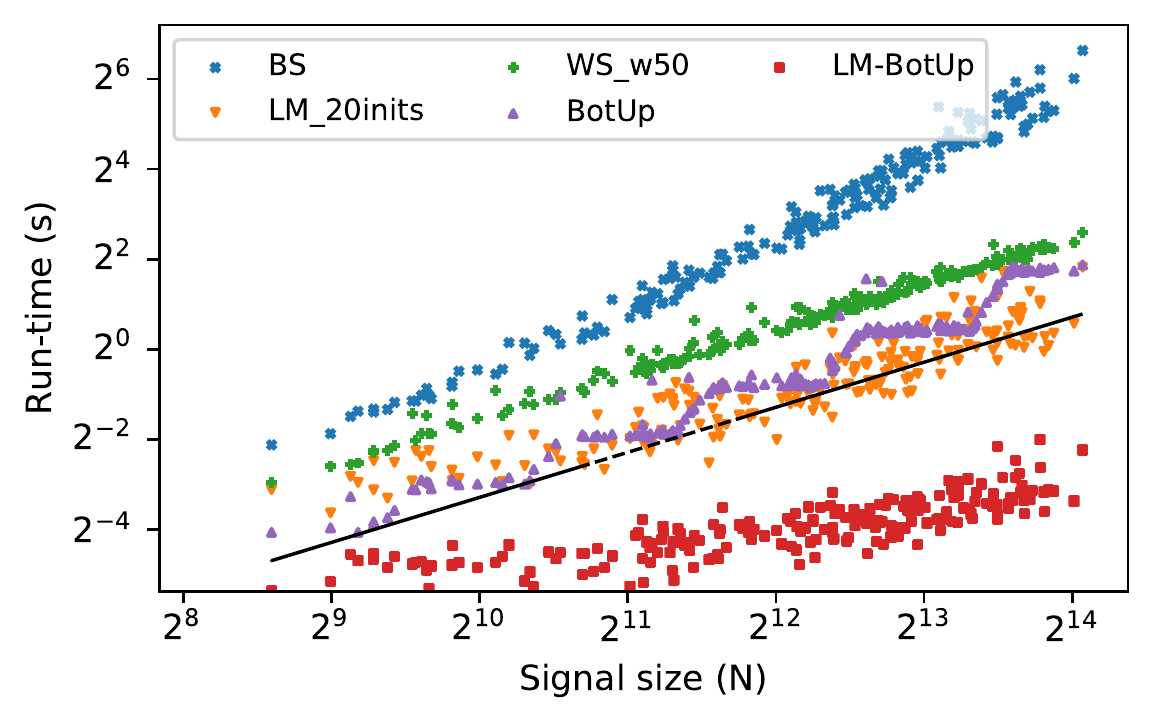}
    \caption{Run-times in seconds (on a single core) for single change point detection, over a range of signal sizes $(N)$. The dashed line shows linear dependence on $N$ for reference.}
  \label{fig:runtime-1-seg} 
\end{figure}

The run-times for each of these algorithms is plotted against signal size $N$ in Figure~\ref{fig:runtime-1-seg}. BS has a steeper slope than 1, because of the $N^2\log(N)$ time complexity. WS\_w50, LM\_20inits, and BotUp follow the expected linearity in signal size, with LM\_20inits and BotUp being faster than WS\_w50. LM-BotUp is the fastest among all of the methods; it runs faster than LM\_20inits because of the fewer iterations involved, and it runs faster than BotUp because of the improved initialisation with fewer cells.

\begin{figure} 
    \centering
    \subfloat{%
        \includegraphics[width=0.95\linewidth]{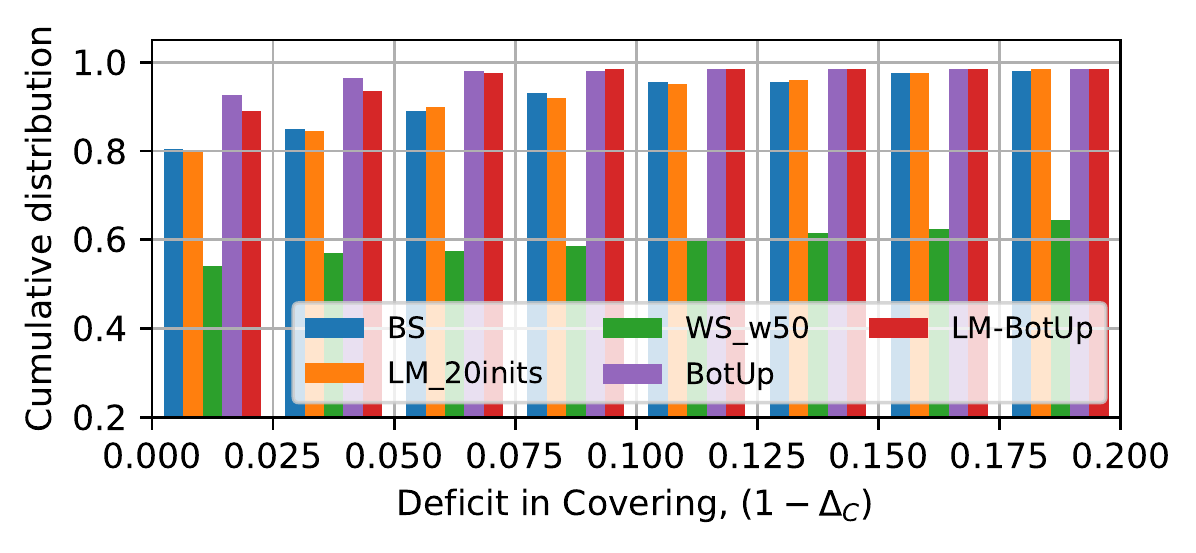}}
        \vspace{0pt}
  \subfloat{%
        \includegraphics[width=0.95\linewidth]{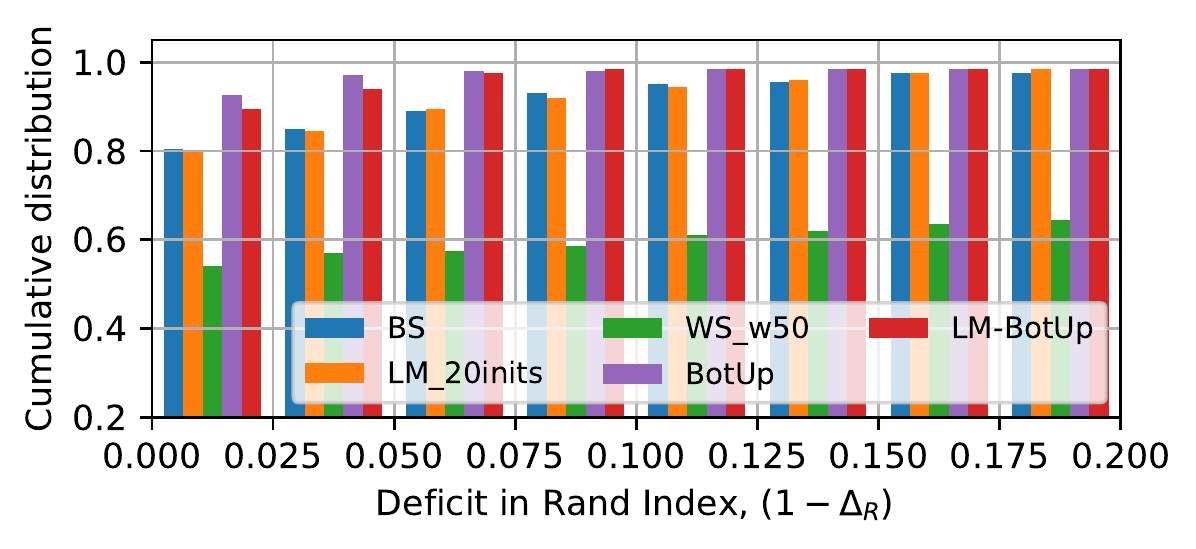}}
    \caption{Accuracy for single change point detection, expressed using a cumulative distribution against deficits, over a synthetic set of 200 signals. Algorithms perform better when the corresponding distribution reaches 1 at a lower deficit. Top: Covering score $\Delta_C$, defined in (\ref{eq:covering}). Bottom: Rand index $\Delta_R$, defined in (\ref{eq:rand-index}).}
  \label{fig:accuracy-1-seg} 
\end{figure}

Figure~\ref{fig:accuracy-1-seg} shows the accuracy for each algorithm as a cumulative distribution over the deficits in covering score and Rand index. BotUp records the best accuracy, with the two metrics being $\geq 0.975$ for over 90\% of the synthetic signals, while LM-BotUp is a close second. LM\_20inits and BS have similar accuracies, while WS\_w50 performs poorly.

\begin{figure} 
    \centering
    \includegraphics[width=0.95\linewidth]{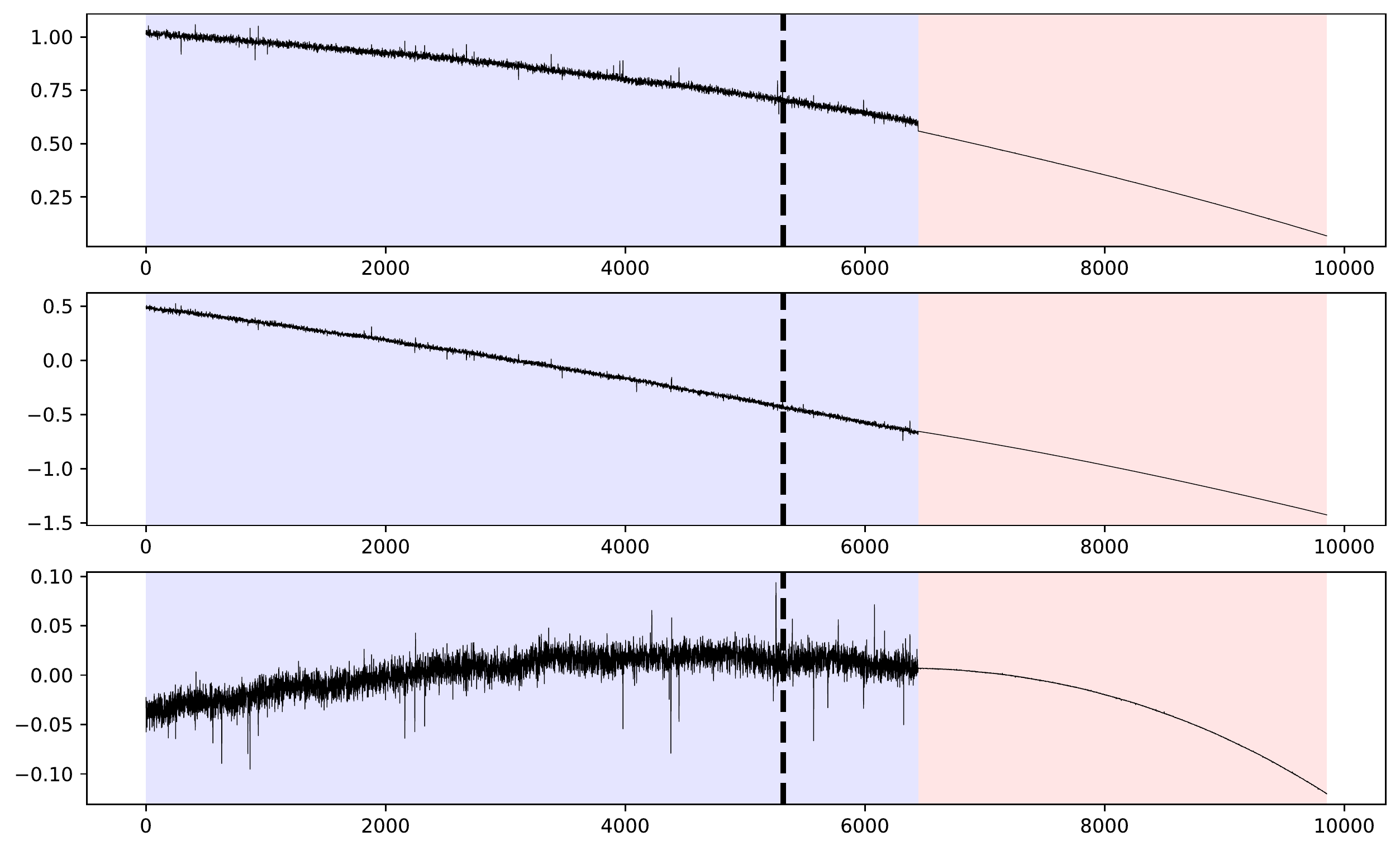}
    \caption{Signal where BotUp shows the greatest improvement in covering score $\Delta_C$ over BS; the first three PCA coordinates are shown from top to bottom. Blue-red shading shows segmentation predicted by BotUp. Vertical dashed line shows change point detected by BS.}
  \label{fig:bs-vs-botup} 
\end{figure}

At first glance, the non-zero deficits for BS can be surprising, since BS finds the optimal change point. And BS does find the optimal change point for each case as dictated by the fitting cost. This can be seen in Table~\ref{tab:avg-1-seg}, where the average fitting costs for each algorithm are shown relative to BS. BotUp and LM-BotUp, which showed the best accuracies \textit{vis-a-vis} the covering score and rand index, have higher fitting costs than BS. This discrepancy in the fitting costs and the accuracy metrics arises because we use weakly non-linear synthetic signals here to simulate real-world datasets. One such signal is illustrated in Figure~\ref{fig:bs-vs-botup}, along with the segmentations identified by BS and BotUp. BS finds the globally optimal change point that minimizes the fitting cost defined in (\ref{eq:cost}). BotUp, instead, has a more localized nature with merging of successive cells. For signals where the parametric model and associated cost function do not adequately capture the complexity of the signal, BotUp is known to provide more accurate predictions \cite{truong2020selective}.

A summary of the performances for different algorithms is provided as average quantities in Table~\ref{tab:avg-1-seg}. LM-BotUp is the fastest, about 6 times faster than BotUp, with accuracies very close to the best case of BotUp. LM\_20inits has accuracies very close to BS at about a tenth of the run-times.

\begin{table}
    \caption{Average performance for $2$-segmentation, with run-times and costs relative to BS.}
    \label{tab:avg-1-seg}
    \centering
    \begin{tabular}{lrrrr}
\toprule
  Algorithm &  Rel. runtime &  Rel. cost &  Covering &  Rand Index \\
\midrule
         BS &         1.000 &      1.000 &     0.978 &       0.979 \\
 LM\_20inits &         0.106 &      1.000 &     0.978 &       0.978 \\
     WS\_w50 &         0.260 &      3.627 &     0.843 &       0.854 \\
      BotUp &         0.118 &      1.027 &     0.989 &       0.990 \\
   LM-BotUp &         0.017 &      1.016 &     0.986 &       0.987 \\
\bottomrule
\end{tabular}

\end{table}

\subsection{Detecting multiple change points over small signals}\label{sec:small-signals}
A set of 200 synthetic signals \textit{containing multiple segments} $k \in \lbrack 2, 10\rbrack$ with sizes $N \in \lbrack 50, 2000\rbrack$, dimensionality $d \in \lbrack 2, 16\rbrack$ are now used to evaluate the performances of the approximate methods, BS, BotUp, LM\_20inits, and LM-BotUp, along with the exact methods, SN, PELT, SNBC\_sf10, and SNBC\_sf50. WS\_w50 is dropped from the comparison because of its poor performance compared to BS and BotUp. A third variant of LM, SNBC\_sf10-LM, is also included. The signal sizes are kept relatively low because SN and PELT computations become expensive for larger sizes.

\begin{figure} 
    \centering
    \includegraphics[width=0.95\linewidth]{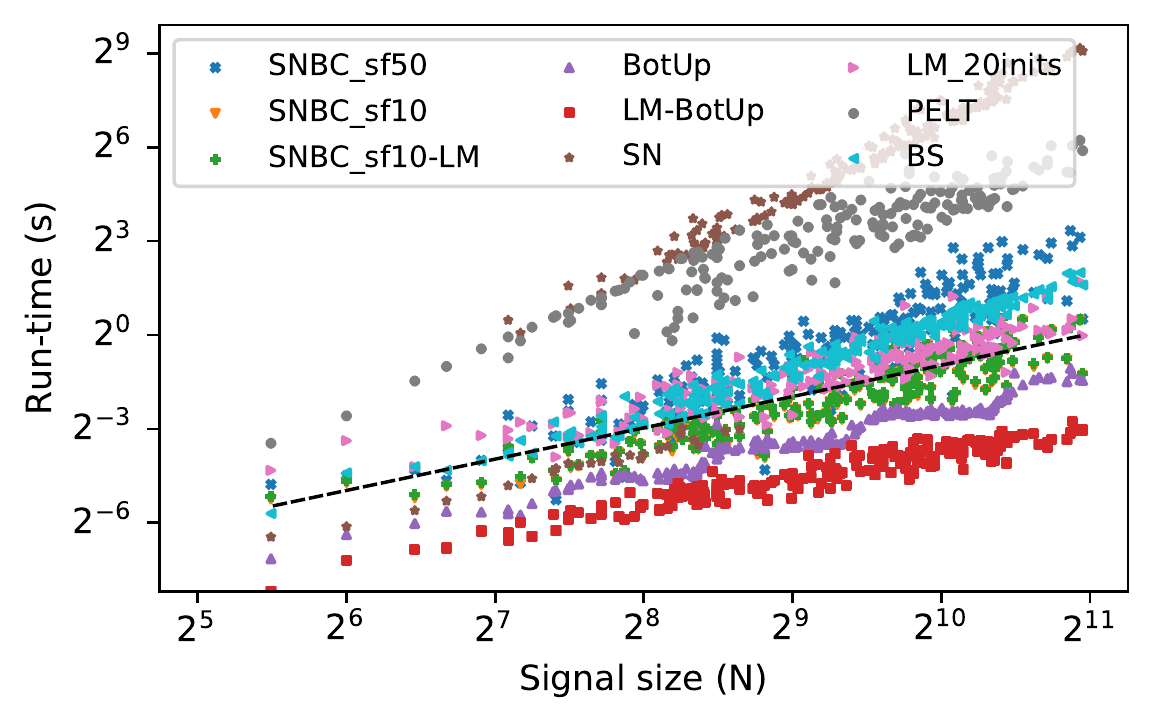}
    \caption{Run-times in seconds (on a single core) for different algorithms for multiple change point detection for small synthetic signals. The dashed line shows linear dependence on $N$ for reference.}
  \label{fig:runtime-Nk-xss} 
\end{figure}

Figure~\ref{fig:runtime-Nk-xss} plots the run-times for each of the algorithms against signal size. Except for BS, SN, and PELT, the other algorithms all show a linear scaling with signal size. This is in line with the estimated time complexities in Table~\ref{tab:algo-complexity}. BotUp and LM-BotUp remain the fastest algorithms as seen previously, with the latter retaining the advantage in speed. 

\begin{figure} 
    \centering
    \subfloat{%
        \includegraphics[width=0.95\linewidth]{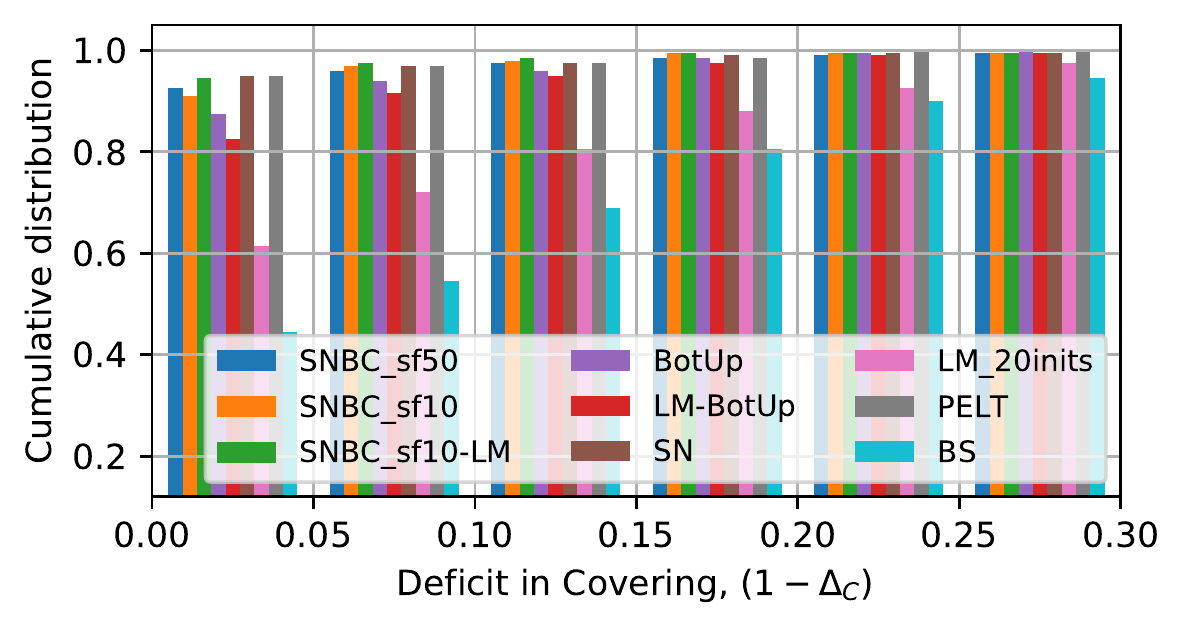}}
    \\
  \subfloat{%
        \includegraphics[width=0.95\linewidth]{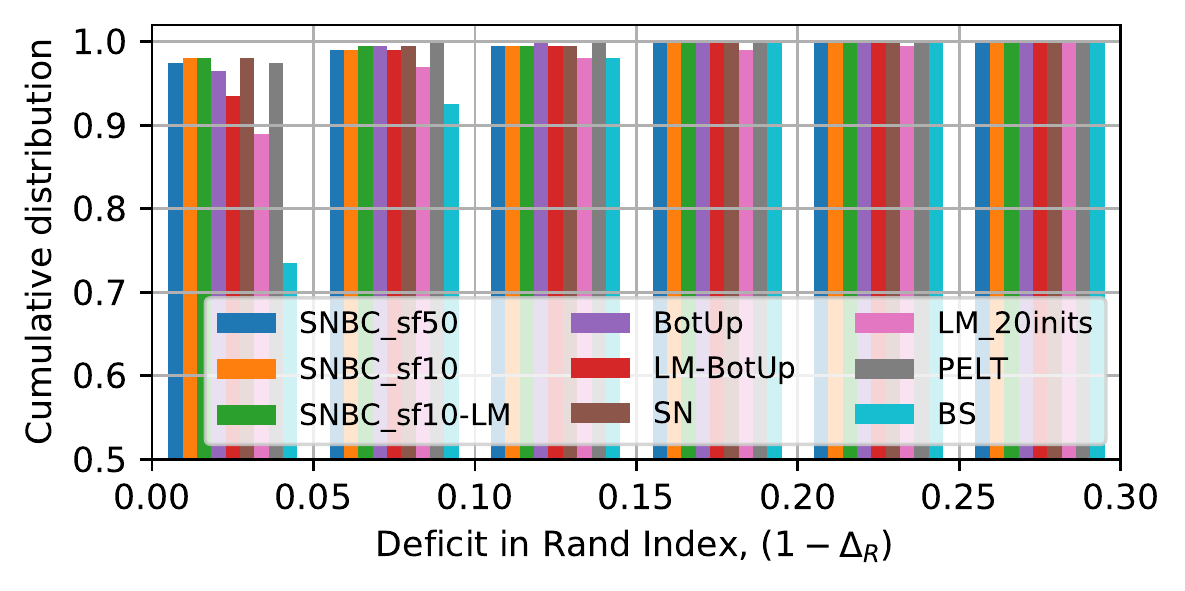}}
    \caption{Accuracy of different $k$-segmentation algorithms for multiple change point detection over small synthetic signals, expressed using a cumulative distribution against deficits over a set of 200 signals. Algorithms perform better when the corresponding distribution reaches 1 at a lower deficit. Top: Covering score $\Delta_C$, defined in (\ref{eq:covering}). Bottom: Rand index $\Delta_R$, defined in (\ref{eq:rand-index}).}
  \label{fig:accuracy-Nk-xss} 
\end{figure}

Figure~\ref{fig:accuracy-Nk-xss} shows accuracies as deficits in covering score and rand index for each of the algorithms. As before, a large value for the cumulative distribution at low deficits indicates better performance. All algorithms except BS and LM\_20 inits show similar performance; although, BotUp and LM-BotUp have slightly fewer cases with scores close to 1. Note that the SNBC\_sf10-LM variant of LM outperforms SNBC\_sf50 and SN in terms of accuracies, with significantly smaller run-times.

A summary in terms of average measures is provided in Table~\ref{tab:avg-Nk-xss}. The run-times and costs are shown relative to the base case of SN. The three variants of LM out-perform the algorithms they modify. SNBC\_sf10-LM is better than SNBC (sf50 and sf10) in accuracy, fitting cost, and run-time. SNBC\_sf10-LM even performs on par with SN while being six times faster. LM-BotUp is very competitive with BotUp in accuracy, and significantly out-performs BotUp in fitting cost and run-time. LM\_20inits outperforms BS in all aspects. Overall, LM-BotUp shows the best performance with competitive accuracies and fitting costs at much lower run-times than the rest.

\begin{table}
    \caption{Average performance over small signals for $k$-segmentation, with run-times and costs relative to SN.}
    \label{tab:avg-Nk-xss}
    \centering
    \begin{tabular}{lrrrr}
\toprule
    Algorithm &  Rel. runtime &  Rel. cost &  Covering &  Rand Index \\
\midrule
    SNBC\_sf50 &         0.349 &      1.182 &     0.979 &       0.995 \\
    SNBC\_sf10 &         0.170 &      1.138 &     0.976 &       0.993 \\
 SNBC\_sf10-LM &         0.177 &      1.066 &     0.983 &       0.997 \\
        BotUp &         0.086 &      2.132 &     0.969 &       0.990 \\
     LM-BotUp &         0.045 &      1.829 &     0.965 &       0.990 \\
           SN &         1.000 &      1.000 &     0.982 &       0.997 \\
   LM\_20inits &         0.360 &      1.501 &     0.929 &       0.981 \\
         PELT &         5.711 &      1.000 &     0.983 &       0.997 \\
           BS &         0.279 &      1.724 &     0.894 &       0.967 \\
\bottomrule
\end{tabular}

\end{table}

\subsection{Detecting multiple change points over large signals}\label{sec:large-signals}
A subset of the previous set of algorithms are now evaluated over a set of 100 synthetic signals. SN, BS, and PELT take much longer than the others; so only SNBC\_sf50, SNBC\_sf10, SNBC\_sf10-LM, BotUp, LM-BotUp, and LM\_20inits are retained. The signals have sizes $N \in \lbrack 4000, 175000\rbrack$, while the dimensionality and number of segments remain the same as before with $d \in \lbrack 2, 16 \rbrack$ and $k \in \lbrack 2, 10\rbrack$ respectively. SNBC\_sf50 is used as the base case to compare run-times and fitting costs.

\begin{figure} 
    \centering
    \includegraphics[width=0.95\linewidth]{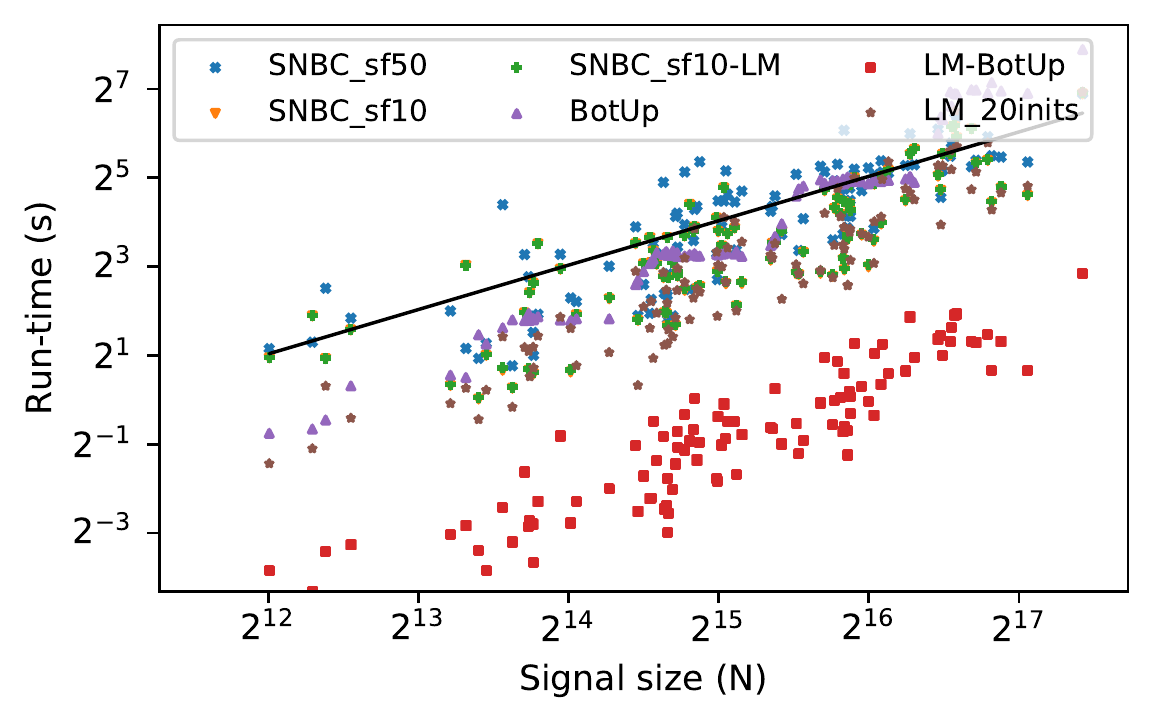}
    \caption{Run-times in seconds (on a single core) for different algorithms for multiple change point detection over large synthetic signals. The dashed line shows linear dependence on $N$ for reference.}
  \label{fig:runtime-Nk-ls} 
\end{figure}

Figure~\ref{fig:runtime-Nk-ls} plots run-times against signal size. As before, linear scaling is observed for all of six algorithms, with LM-BotUp remaining the fastest among them.

\begin{figure} 
    \centering
    \subfloat{%
        \includegraphics[width=0.95\linewidth]{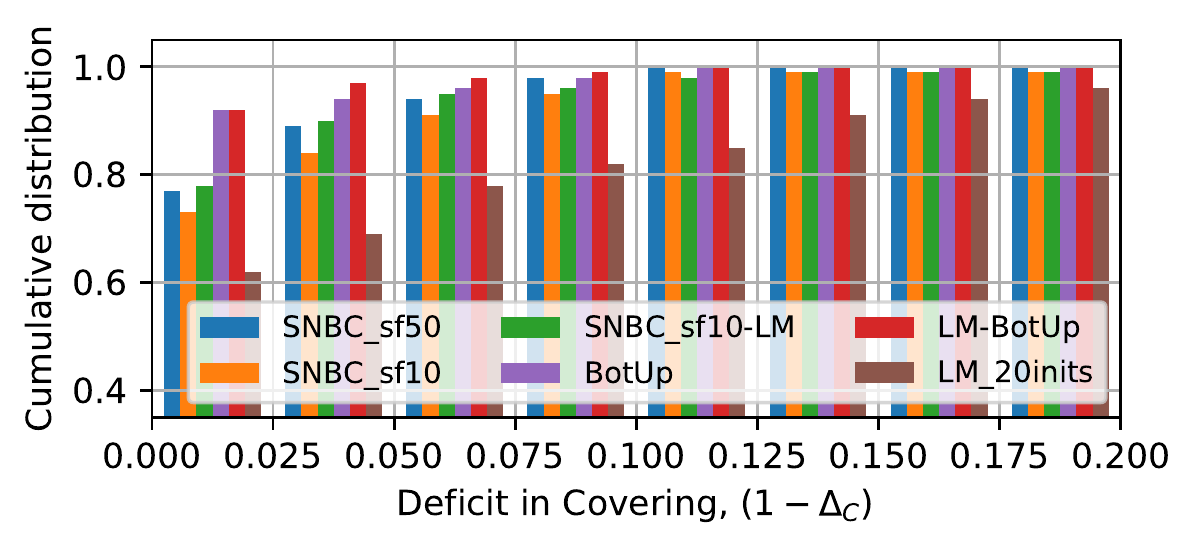}}
    \\
  \subfloat{%
        \includegraphics[width=0.95\linewidth]{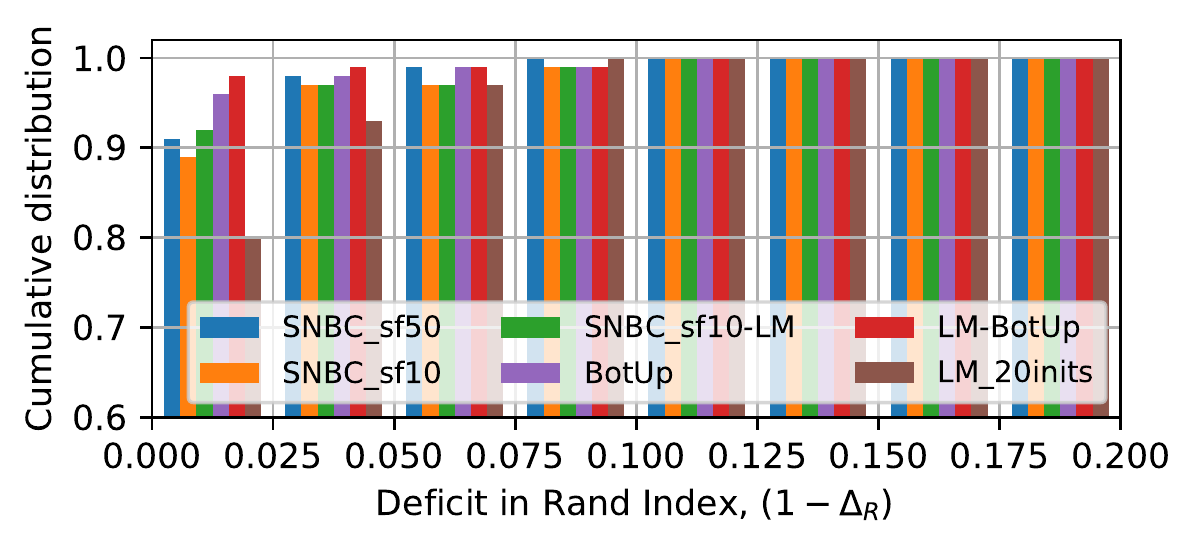}}
    \caption{Accuracy of different $k$-segmentation algorithms for multiple change point detection over large synthetic signals, expressed using a cumulative distribution against deficits over a set of 100 signals. Algorithms perform better when the corresponding distribution reaches 1 at a lower deficit. Top: Covering score $\Delta_C$, defined in (\ref{eq:covering}). Bottom: Rand index $\Delta_R$, defined in (\ref{eq:rand-index}).}
  \label{fig:accuracy-Nk-ls} 
\end{figure}

Figure~\ref{fig:accuracy-Nk-ls} shows accuracies in terms of deficits in covering score and Rand index. A summary in terms of average measures is provided in Table~\ref{tab:avg-Nk-ls}. The results remain consistent with those in Section~\ref{sec:small-signals}, with the LM-variants outperforming the algorithms they modify. In fact, LM-BotUp comes out as the clear winner over BotUp for the large signals considered here. LM-BotUp has better accuracies and fitting costs, but at run-times that are only about 4\% of those of BotUp and SNBC-variants. As noted earlier, SN and PELT take much longer than even SNBC\_sf50.

\begin{table}
    \caption{Average performance over large signals for $k$-segmentation, with run-times and costs relative to SNBC\_sf50.}
    \label{tab:avg-Nk-ls}
    \centering
    \begin{tabular}{lrrrr}
\toprule
    Algorithm &  Rel. runtime &  Rel. cost &  Covering &  Rand Index \\
\midrule
    SNBC\_sf50 &         1.000 &      1.000 &     0.983 &       0.992 \\
    SNBC\_sf10 &         0.885 &      1.026 &     0.976 &       0.989 \\
 SNBC\_sf10-LM &         0.897 &      0.995 &     0.982 &       0.992 \\
        BotUp &         1.202 &      1.017 &     0.992 &       0.996 \\
     LM-BotUp &         0.042 &      1.006 &     0.993 &       0.997 \\
   LM\_20inits &         0.538 &      1.194 &     0.956 &       0.986 \\
\bottomrule
\end{tabular}

\end{table}

\subsection{$7$-segmentation for BDD100k videos}\label{sec:bdd}
Five algorithms --- SNBC\_sf50, SNBC\_sf10, SNBC\_sf10-LM, BotUp, and LM-BotUp --- are used to compute $7$-segmentations for a sample set of 200 scenes from BDD100k \cite{yu2020bdd100k}; the choice of $k=7$ is arbitrarily chosen for illustration. These videos are first processed using a MobileNet featurizer, and the first 32 PCA projections of the 1280-dimensional features are used for $7$-segmentation. The average run-times and relative fitting costs are shown in Table~\ref{tab:avg-bdd}. Note that the run-times are now reported in seconds because all signals have the same size ($N\approx 1200$) and dimensionality ($d=32$ after PCA truncation). The fitting cost is shown relative to SNBC\_sf50.

As with earlier experiments, the two LM variants improve upon the algorithms they modify. SNBC\_sf10-LM has a lower fitting cost than SNBC\_sf10 for a small increase in run-time; SNBC\_sf10-LM is over four times faster than SNBC\_sf50 for only a 0.3\% increase in fitting cost. LM-BotUp has the same fitting cost while being faster than BotUp. Note that the speed-up of LM-BotUp becomes much more pronounced for large signals as seen in Section~\ref{sec:large-signals}.

Covering score and Rand index are not reported here because, unlike the synthetic signals, scenes in BDD100k do not come with the ground truth for $7$-segmentation. The segmented videos due to four of these algorithms (excluding SNBC\_sf10) are provided in \cite{supplementary2020} to demonstrate $k$-segmentation, and for visual comparison of the algorithms; only four algorithms are shown so that the videos can be tiled into one. The segmented videos are reconstructed at a reduced resolution to restrict file sizes. Segmentations are shown using colored borders that change from one segment to the next.

\begin{table}
    \caption{Average performance for $7$-segmentation over 200 sample scenes from BDD100k.}
    \label{tab:avg-bdd}
    \centering
    \begin{tabular}{lrr}
\toprule
    Algorithm &  Runtime (s) &  Rel. cost \\
\midrule
    SNBC\_sf50 &        2.682 &      1.000 \\
    SNBC\_sf10 &        0.574 &      1.013 \\
 SNBC\_sf10-LM &        0.588 &      1.003 \\
        BotUp &        0.159 &      1.052 \\
     LM-BotUp &        0.096 &      1.052 \\
\bottomrule
\end{tabular}

\end{table}

\subsection{Other considerations}
The computational cost of $k$-segmentation is often reduced by decreasing the sampling rate; i.e., by selecting every $m^{th}$ ($m>1$) point in the signal for analysis. Such down-sampling can be used for any of the algorithms discussed here, and therefore does not affect the relative performances of these algorithms. The advantages of LM-enhanced algorithms remain considerable even when down-sampling is used.

The $k$-segmentation problem was framed as one where the number of segments $k$ is known. In practice, this is often not the case. The new LM algorithm is easily extensible to the $k$-unknown $k$-segmentation problem by incorporating an $L_0$-penalty term, as is commonly done for such problems. The LM algorithm is also easily applied to online learning problems. A popular choice for online $k$-segmentation is to compare $2$-segmentation against $1$-segmentation over a specified window size that is split into equal parts. With LM, these window sizes can be made significantly larger, and the number of segmentations can be drastically reduced due to the heuristic nature of the algorithm.

\section{Conclusion}\label{sec:conclusion}
The $k$-segmentation of videos can provide an efficient way to explore, summarize, and select data for training machine learning models. The same $k$-segmentation framework is also applicable to other domains such as finance, economics, and bioinformatics, which involve problems where time-indexed data needs to be separated into disjoint partitions. A new heuristic algorithm called the LM algorithm has been introduced in this paper. It has linear time complexity, and can take any initial guess for candidate change points to produce a locally optimal set of change points. The cheap computational cost paired with its ability to operate over any initial guess makes this algorithm very flexible. It is particularly effective in accelerating existing algorithms.

A variant of the new algorithm, LM-BotUp, uses LM for the initial steps before using Bottom-Up segmentation to find the required $k$-segmentation. This algorithm is often competitive or better than all existing algorithms in terms of accuracy and fitting costs, but at run-times that can be as small as 4\% of the best among the rest for datasets with $\sim 50000$ points. Even for hour-long-videos (assuming image featurization is done), LM-BotUp can compute $k$-segmentations in only a few seconds on a single CPU core.

\bibliographystyle{IEEEtran}
\bibliography{biblio}
\end{document}